%% file: content.tex
\documentclass[10pt,twocolumn,letterpaper]{article}

\usepackage{cvpr}

\usepackage{abbrv}
\usepackage[dvipsnames]{xcolor}
\usepackage{url}
\usepackage{tikz}
\usepackage{comment}
\usepackage{amsmath,amssymb,bbm}
\usepackage{colortbl}
\usepackage{epsfig}
\usepackage{caption}
\usepackage{soul}
\usepackage[utf8]{inputenc}
\usepackage{listings}
\usepackage[T1]{fontenc}
\usepackage{booktabs}
\usepackage{amsfonts}
\usepackage{nicefrac}
\usepackage{microtype}
\usepackage[noend]{algorithmic-mine}
\usepackage{algorithm-mine}
\usepackage{numprint}
\usepackage{siunitx}
\usepackage{etoolbox}
\usepackage{cancel}
\newcommand{\ubold}{\fontseries{b}\selectfont}
\robustify\ubold
\usepackage{wrapfig}
\usepackage[skip=2pt,font=footnotesize,labelfont=footnotesize]{subcaption}
\usepackage{pifont}
\usepackage{lineno}
\usepackage{array,multirow,graphicx}
\usepackage[para]{footmisc}
\usepackage{lipsum}  
\usepackage{stfloats}
\usepackage[colorinlistoftodos,prependcaption,textsize=tiny]{todonotes}

\usepackage[accsupp]{axessibility}
\usepackage{pgfplots}

\pgfplotsset{compat=1.18}

\definecolor{cvprblue}{rgb}{0.21,0.49,0.74}
\usepackage[pagebackref,breaklinks,colorlinks,allcolors=cvprblue]{hyperref}
\usepackage{fontawesome}

\definecolor{keyword}{rgb}{.224,.451,.686}

\newcommand{\cmark}{\textcolor{OliveGreen}{\ding{52}}}
\newcommand{\xmark}{\textcolor{BrickRed}{\ding{56}}}

\newcommand{\keyword}[1]{\textcolor{keyword}{#1}}
\newcommand{\keywordone}[1]{\textcolor{Red}{#1}}
\newcommand{\keywordtwo}[1]{\textcolor{Green}{#1}}
\newcommand{\keywordtri}[1]{\textcolor{Purple}{#1}}

\newcommand{\alname}{InsTALL}

\def\paperID{13601}
\def\confName{CVPR}
\def\confYear{2025}

\title{\alname: Context-aware Instructional Task Assistance with Multi-modal\\Large Language Models}

\author{Pha Nguyen\textsuperscript{*, \faAmazon} \quad Sailik Sengupta\textsuperscript{\faAmazon} \quad Girik Malik\textsuperscript{\faAmazon} \quad Arshit Gupta\textsuperscript{\faAmazon} \quad Bonan Min\textsuperscript{\faAmazon}\\
\small \textsuperscript{*} University of Arkansas \qquad \textsuperscript{\faAmazon} AWS AI Labs \\
\small \textsuperscript{*} {\tt panguyen@uark.edu} \quad \textsuperscript{\faAmazon} {\tt \{sailiks, girikm, arshig, bonanmin\}@amazon.com}
}

\begin{document}
\input{main}

{
\small
\bibliographystyle{ieeenat_fullname}
\bibliography{egbib}
}

\clearpage

\renewcommand{\thesection}{\Alph{section}}
\renewcommand{\thetable}{\Alph{section}.\arabic{table}}
\renewcommand{\theequation}{\Alph{section}.\arabic{equation}}
\renewcommand{\thealgorithm}{\Alph{section}.\arabic{algorithm}}

\end{document}

%% file: main.tex
\twocolumn[{
 \renewcommand\twocolumn[1][]{#1}%
 \maketitle
 \begin{center}
 \centering
 \captionsetup{type=figure}
 \includegraphics[width=\textwidth,trim={0 65cm 0 62cm},clip]{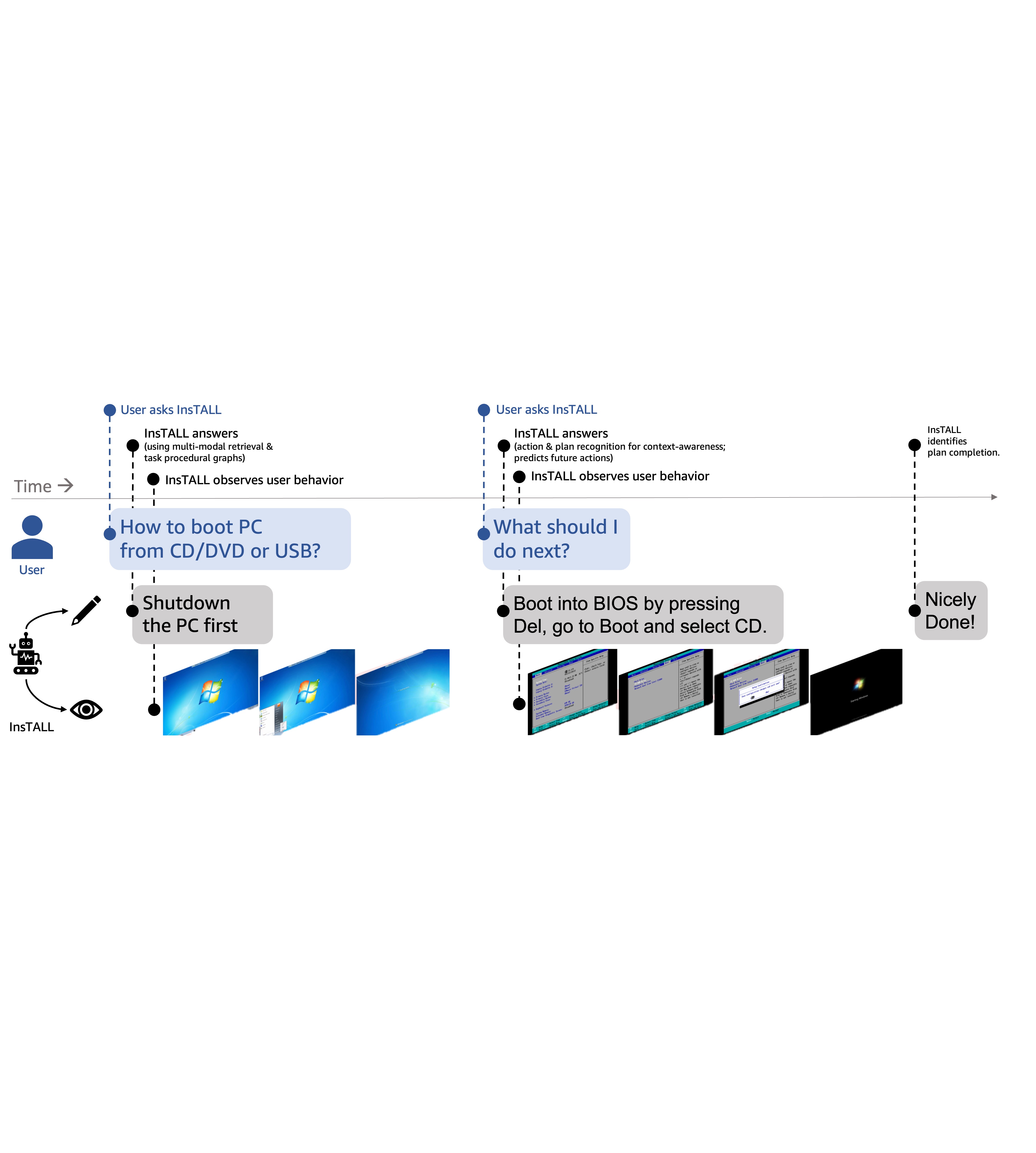}
 \captionof{figure}{\alname~showcasing its ability to understand visual cues of the user's environment and comprehend user instructions to provide context-aware assistance.}
 \label{fig:teaser}
 \end{center}
 }]

\begin{abstract}%
The improved competence of generative models can help building multi-modal virtual assistants that leverage modalities beyond language. By observing humans performing multi-step tasks, one can build assistants that have situational awareness of actions and tasks being performed, enabling them to cater assistance based on this understanding. In this paper, we develop a Context-aware Instructional Task Assistant with Multi-modal Large Language Models (\alname) that leverages an online visual stream (\eg a user's screen share or video recording) and responds in real-time to user queries related to the task at hand. To enable useful assistance, \alname~1) trains a multi-modal model on task videos and paired textual data, and 2) automatically extracts task graph from video data and leverages it at inference time. We show \alname~
achieves state-of-the-art performance across proposed sub-tasks considered for multimodal activity understanding-- task recognition (TR), action recognition (AR), next action prediction (AP), and plan prediction (PP)-- and outperforms existing baselines on two novel sub-tasks related to automatic error identification.
\end{abstract}
\noindent\rule{0.48\textwidth}{0.2pt}
{\footnotesize \textsuperscript{*} Work done as an intern at Amazon.}

\section{Introduction}%
\label{sec:intro}

In recent years, Multimodal Large Language Models (MLLMs) have shown remarkable advancements in various multi-modal tasks~\cite{wu2024nextgpt}. For example, vision-language models have achieved significant success in areas such as visual captioning and visual question answering~\cite{li2023blip, liu2024improved}. In essence, these models have demonstrated the ability to understand the visual content in images while being able to follow human instructions. To do so, these works often use a lightweight adapter that connects a visual encoder to a language model and pre-trains the composite network on large-scale multi-modal datasets, at times followed by fine-tuning on task-specific datasets for downstream applications. Beyond images, researchers have further extended the capabilities of MLLMs to consider procedural tasks in videos~(VideoLLM~\cite{chen2024videollm}; \cite{ashutosh2024detours}). With the widespread availability of instructional videos that demonstrate multistep tasks~\cite{malmaud-etal-2015-whats, alayrac2016unsupervised, zhou2018towards, sener2022assembly101, zhukov2019cross, tang2019coin, miech2019howto100m}, there is an opportunity to develop systems that can understand the actions that are being performed in the context of a task and provide context-aware assistance. In the paper, we seek to empower MLLMs to answer real-time user queries related to various sub-tasks related to this goal, such as Task Recognition (TR), Action Recognition (AR), Next Action Prediction (AP), and Plan Prediction (PP).

Prior works have shown that encoding visual tokens into LLMs' input space, using a visual encoder and translator, can aid in the seamless use of visual and text signals \cite{chen2024videollm}. In online settings, leveraging such an architecture alongside real-time video frames has also been shown to improve performance across the aforementioned sub-problems \cite{ashutosh2024video}. Interestingly, these works train models on extensive dialog or narration data alongside video inputs and over rely on the generalization capability of LLMs to make sense of action dependencies in tasks. Given that previous works have shown the complexity of assistance for planning tasks \cite{grover2020radar} and questionable planning capabilities of current LLMs~\cite{valmeekam2023planning}, we revisit this assumption and seek to improve the performance of MLLM-based online assistants for procedural tasks.

\input{tables/all_related_works_table}

\paragraph{Contribution.} In this work, we enable more effective and contextual guidance for users engaged in multi-step tasks by introducing several key innovations. First, we leverage the VideoLLM style architecture (with Mistral as the base LLMs) and introduce inductive bias by developing query prompts for narration, planning, answering questions, and detecting mistakes; this minimizes the need for extensive manual annotation (Table~\ref{tab:prompt_types}). Second, we leverage existing video data to understand task dependencies and construct graph representations that empower reliable guidance across sub-tasks necessary for offline Task Recognition (Eqn.~\eqref{eq:ar_graph}) and online (Eqn.~\eqref{eq:sr_graph},~\eqref{eq:sa_graph},~\eqref{eq:pa_graph},~\eqref{eq:pap_graph}) conversational guidance based on contextual video understanding (\S\ref{subsec:context}). We note that \alname~is, to the best of our knowledge, the first to enable MLLMs to learn from both graphical and visual representations (Fig.~\ref{fig:framework}) and enable the use of both VectorRAG~\cite{lewis2020retrieval, izacard-grave-2021-leveraging, guu2020retrieval, gao2023retrieval, levy2024chatting} and GraphRAG~\cite{edge2024local, xu2024retrieval, chen-etal-2020-kgpt} style approaches for real-time multi-modal assistance. Finally, our comprehensive evaluation on a holistic set of five sub-tasks across two datasets showcases that \alname~consistently outperforms existing state-of-the-art (SoTA) methods and closed-source models.

\section{Related Work}
\label{sec:formatting}

\input{tables/task_coverage_table}

\paragraph{Multimodal Large Language Models}
Prior works on Multi-modal LLMs consider using pre-trained encoders to transform images onto an LLM's input token space
~\cite{huang2023language, zhu2024minigpt, su2022language, koh2024generating}. Some works improve upon this base multi-modal encoding mechanism, such as Flamingo~\cite{alayrac2022flamingo} which uses a multi-modal cross-attention mechanism across all layers, while others like BLIP-2~\cite{li2023blip} incorporate a lightweight transformer model to merge image and text before the LLM input stage. Subsequently, works have adopted similar practices for other modalities, such as video~\cite{song2024moviechat, zhang-etal-2023-video, he2024ma} and audio~\cite{zhang-etal-2023-speechgpt}. PandaGPT~\cite{su-etal-2023-pandagpt} builds upon this and is able to comprehend six different modalities simultaneously by integrating a multimodal encoder~\cite{girdhar2023imagebind}. The improvements have empowered recent works to explore multi-modal decision-making problems~\cite{wu2023visual, shen2024hugginggpt, huang2024audiogpt}.

\paragraph{Instructional Video Understanding.}
Beyond fully autonomous computer usage \cite{Anthropic_2024} or autonomous driving scenarios, lies a crucial realm of assistance that requires a contextual understanding of visual cues and temporal grounding \cite{lan2023survey, zhang2023temporal}. In such cases, the agent takes as input a video alongside a textual query and seeks to provide assistance in textual format. These videos may belong to various domains, such as cooking~\cite{afouras2024ht, regneri2013grounding}, daily activities~\cite{caba2015activitynet}, indoor scenes~\cite{gao2017tall}, and movies~\cite{lei2020tvr}. Previous approaches have relied on sliding window-based methods~\cite{anne2017localizing, gao2017tall, liu2018attentive} and scanning-and-ranking-based techniques~\cite{xu2019multilevel, chen2019semantic, ge2019mac, chen2018temporally, zhang2019man, liu2020jointly} for visual understanding. The leveraged video understanding can then be interpolated into the text space to identify actions and enable procedure/task planning based on the textual predicates/states and video frames \cite{chang2020procedure, bi2021procedure, sun2022plate,zhao2022p3iv, wang2023pdpp, fang2023masked, li2023skip, wang2023event, niu2024schema} (where the latter works have leveraged diffusion ~\cite{ho2020denoising} and/or transformer~\cite{vaswani2017attention} models).
Recent developments in MLLMs reformulate the problem of visual question answering with online video clips~\cite{chen2024videollm} by relying on the reasoning capabilities of the backbone LLMs. Recent works have also critiqued the planning capabilities of LLMs \cite{valmeekam2023planning}. In this paper, we improve the planning abilities of MLLMs to support instructional assistance tasks by leveraging Task Procedural Graphs. 

\paragraph{Procedural Graphs (PG)}
In this regard, PGs offer a structured representation of the sequential steps and transitions involved in a given task. Recent works highlight that utilization of graph structures in LLMs can enhance semantic understanding and reasoning capabilities~\cite{lin2024graphenhanced, sakaguchi-etal-2021-proscript-partially}. Further, automatically generating plausible plans for daily tasks shows that LLMs can be used to develop reasonable ordering of actions and goal identification~\cite{madaan-etal-2022-language, xie2023translating, yuan-etal-2023-distilling}, while others have expressed a limit to their effectiveness \cite{roy-etal-2024-flap}.
In essence, incorporating this graph-based knowledge helps models to better comprehend the logical flow and the dependencies between steps. We take advantage of this idea for MLLMs and showcase its efficacy in improving instructional assistance tasks. \autoref{tab:features} compares key features of various approaches to understanding instructional videos and shows that our approach, \alname, uniquely leverages \textbf{PG} alongside other input signals. Further, \alname~ supports video retrieval (\textbf{Retr.}) that draws inspiration from the notion of GraphRAG \cite{edge2024local} but for multi-modal RAG scenarios. In addition, \autoref{tab:prompt_types} reports that the number of annotated samples constructed using PG (\S\ref{leverage_pg}); it is much larger than recent MLLM approaches~\cite{chen2024videollm}.

\subsection{Discussion}

Our approach aims to provide a more comprehensive and interactive experience for the instructional assistant. Specifically, we \textit{(i)} formally model the procedures involved in multi-step tasks (Alg.~\ref{alg:graph_construction}) and \textit{(ii)} generalize this knowledge into a representation to support the assistant's understanding (Eqn.~\eqref{eq:MLLM_with_G}). Furthermore, contextual awareness enables the assistant to \textit{(iii)} flexibly train on different objectives for the language model (Eqn.~\eqref{eq:tr},~\eqref{eq:ar},~\eqref{eq:ap}, and~\eqref{eq:pp}). Moreover, our approach \textit{(iv)} diversifies the user's queries, creating an online streaming dialog that simulates a natural conversation (\cref{subsec:context} and~\cref{subsec:mistake}). This is a significant advance over previous work that has focused mainly on annotations for single-shot question-answering~\cite{yu2019activitynet}. Through this comprehensive approach, our aim is to develop an instructional assistant who not only \textit{understands the procedures and knowledge involved} but also provides \textit{interactive assistance} tailored to the needs of the user and the state of the task at hand. Note that our approach is a multimodal LLM-based technique that uniquely employs a procedural graph to {model context and enhance recognition and forecasting capabilities}. It is important to note that the predictions are not simply derived from the graph mining process used in previous works~\cite{ashutosh2024video, liu2020jointly, zhou2023procedure}.

\section{Objectives for Multi-task Learning}\label{sec:prob_formu}

We seek to design a single Multi-modal LLM (MLLM) that is capable of performing well on several sub-tasks necessary for clear instructional assistance. To achieve this, we define a prompt $\mathbf{Q}_{\text{task}}$ for each task that enables the MLLM to adapt its behavior and outputs based on the assistance scenario at hand. We denote $\mathbf{V} = \{\mathbf{v}_t \ | \ 0 \leq t < |\mathbf{V}|\}$ as the video associated with a particular activity or task $\mathbf{T}$ (\eg \texttt{cooking omelette}), where $\mathbf{v}_t$ are \textit{action clips} denoting an action $\textbf{a}_{t}$ (\eg \texttt{fry eggs}) used to perform the task. Now, we describe four tasks:

\paragraph{Task Recognition (TR)} Given a video snippet $\mathbf{V}$ and a task prompt $\mathbf{Q}_{\texttt{TR}}$, we seek to identify the task being performed by minimizing the following objective:
\begin{equation}
\min \mathbb{E}_{\mathbf{V}, Y} \bigg[-\sum_{i=1}^{n} Y_i\log\Big(\mathbbm{1}_{Y}\big(p(\mathbf{T} | \mathbf{V}, \mathbf{Q}_{\texttt{TR}})\big)_i\Big)\bigg]
\label{eq:tr}
\end{equation}
where $\mathbf{T}$ is the text response from the MLLM. As this is a classification task, we expect a one-hot mapping that maps the response to the set of task categories $Y$, \ie, denoted as $\mathbbm{1}_{Y}(\cdot)$, and $n = |Y|$.

\paragraph{Action Recognition (AR)}
Given a clipped video $\mathbf{v}_{t}$ and a task prompt $\mathbf{Q}_{\texttt{AR}}$, we seek to identify the action being performed in it by minimizing the following objective:
\begin{equation}
\min \mathbb{E}_{\keyword{\mathbf{v}_{t}}, y} \bigg[-\sum_{i=1}^{m} y_i\log\Big(\mathbbm{1}_{y}\big(p(\mathbf{a}_{t} | \keyword{\mathbf{v}_{t}}, \mathbf{Q}_{\texttt{AR}})\big)_i\Big)\bigg]
\label{eq:ar}
\end{equation}
where $\mathbf{a}_{t}$ is the answer and $y_i$ is the action/step annotation for clip $\mathbf{v}_{t} (\in \mathbf{V})$, and $m = |y|$.

\paragraph{Action Prediction (AP)}
Given the task prompt $\mathbf{Q}_{\texttt{AP}}$, a video upto a particular point $\mathbf{v}_{<t}$, we learn to predict the next likely step $\mathbf{a}_{t}$ by minimizing the objective below:
\begin{equation}
\min \mathbb{E}_{\keyword{\mathbf{v}_{< t}}, y} \bigg[-\sum_{i=1}^{m} y_i\log\Big(\mathbbm{1}_{y}\big(p(\mathbf{a}_{t} | \keyword{\mathbf{v}_{< t}}, \mathbf{Q}_{\texttt{AP}})\big)_i\Big)\bigg]
\label{eq:ap}
\end{equation}

\paragraph{Plan Prediction (PP)}
Given the task prompt $\mathbf{Q}_{\texttt{PP}}$, a video upto a particular point $\mathbf{v}_{<t}$, we seek to predict an ordered list of actions $\mathbf{a}_{\geq t}$ by minimizing the multiple-class mapping function $\mathbb{T}_{y}(\cdot)$:
\begin{equation}
\min \mathbb{E}_{\keyword{\mathbf{v}_{< t}}, y} \bigg[-\sum_{i=1}^{m} y_i\log\Big(\keyword{\mathbb{T}_{y}}\big(p(\keyword{\mathbf{a}_{\geq t}} | \keyword{\mathbf{v}_{< t}}, \mathbf{Q}_{\texttt{PP}})\big)_i\Big)\bigg]
\label{eq:pp}
\end{equation}
where the number of procedural steps in $|\keyword{\mathbf{a}_{\geq t}}| > |\mathbf{a}_t| = 1$.

With all the task objectives defined, we now explore how to teach LLMs all these objectives and incorporate additional knowledge from a procedural knowledge graph.

\section{Developing \alname}

In this section, we relax the assumption imposed by prior work on developing online video assistance \cite{chen2024videollm}; namely, its reliance on the dependency understanding capabilities of LLMs for procedural tasks. Specifically, we investigate how the integration of procedural graphs can be used to generate contextually accurate responses for the various tasks.

\subsection{Designing Multimodal LLM (MLLMs)}

\begin{figure}[!t]
 \centering
 \includegraphics[width=0.48\textwidth]{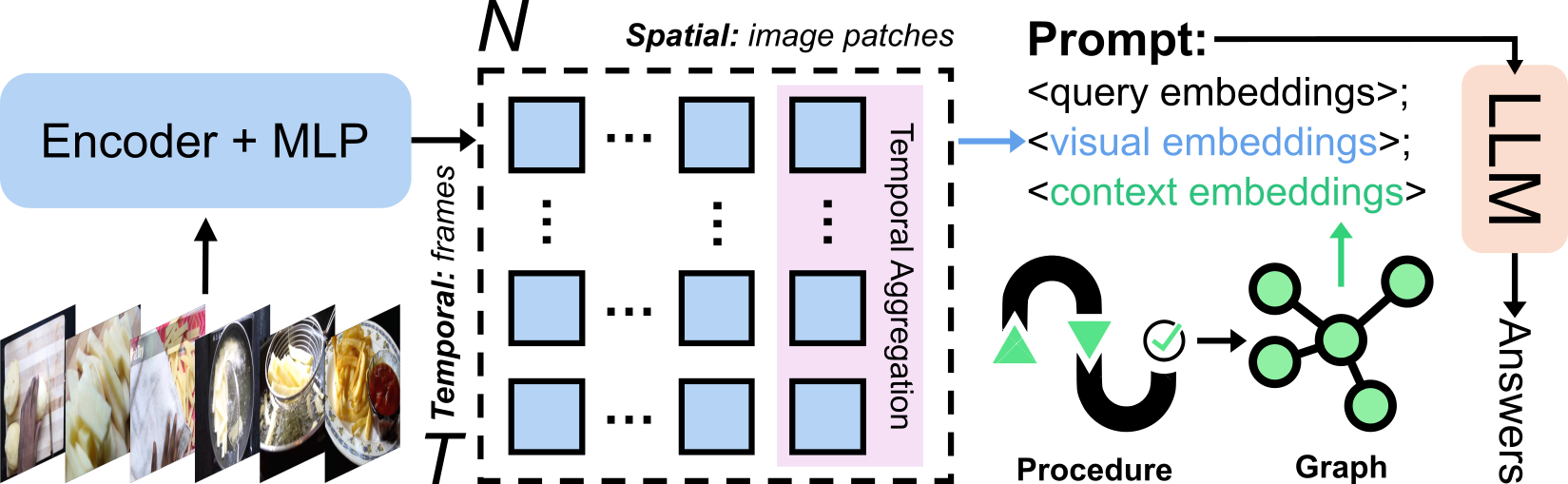}
 \vspace{0.5em}
 \caption{\alname~comprises of an image encoder, an MLP projector, a temporal aggregator, and an LLM. An input sequence of video frames is processed by the image encoder followed by the MLP. The extracted spatio-temporal features are shown using a grid of image patches across multiple frames, where each frame embedding has $N$ pooled spatial tokens. We then compress $T\times N$ embeddings along the temporal axis. The \textit{MLP} helps transform these video embeddings to the text space. In addition, \alname~includes a graph structure constructed from task procedures and language tokens, all input to the LLM.}
 \label{fig:framework}
\end{figure}

Our model takes as input a video content $\mathbf{V}$ and a query $\mathbf{Q}$, and auto-regressively generates a text response of length $L$ denoted as the target answer $\mathbf{A} = [x_0, \dots, x_i, \dots, x_{L-1}]$.
\begin{equation}
p(\mathbf{A} | \keyword{\mathbf{V}}, \mathbf{Q}) = \prod_{i=0}^{L-1} p(x_i | \keyword{\mathbf{V}}, \mathbf{Q}, x_{<i})
\end{equation}%
The model architecture, shown in Fig.~\ref{fig:framework}, is similar to LLaVA~\cite{liu2024visual}. It comprises of an image encoder, a temporal aggregator, a Multi-Layer Perceptron (MLP) layer, and a language model. For the image encoder, we utilize CLIP ViT-L~\cite{radford2021learning, dosovitskiy2021an} to extract embeddings for each video frame.  Then, the model extracts spatio-temporal features using a grid of image patches across multiple frames. Each frame embedding has $N$ pooled spatial tokens where a \textit{temporal aggregator} compresses $T\times N$ embeddings along the temporal axis. The resulting video embeddings from the temporal aggregator are then projected using an MLP to frame tokens that are then interleaved with language tokens as input to a large language model. In our experiments, we consider the Mistral-7B-Instruct \cite{jiang2023mistral} as the language model. Finally, we add  LoRA~\cite{hu2022lora} parameters with every linear layer of the language model for efficient learning of the tasks in \S\ref{sec:prob_formu}.

\subsection{Leveraging Procedural Graph}\label{leverage_pg}

\begin{figure}[!t]
 \centering
 \includegraphics[width=0.48\textwidth]{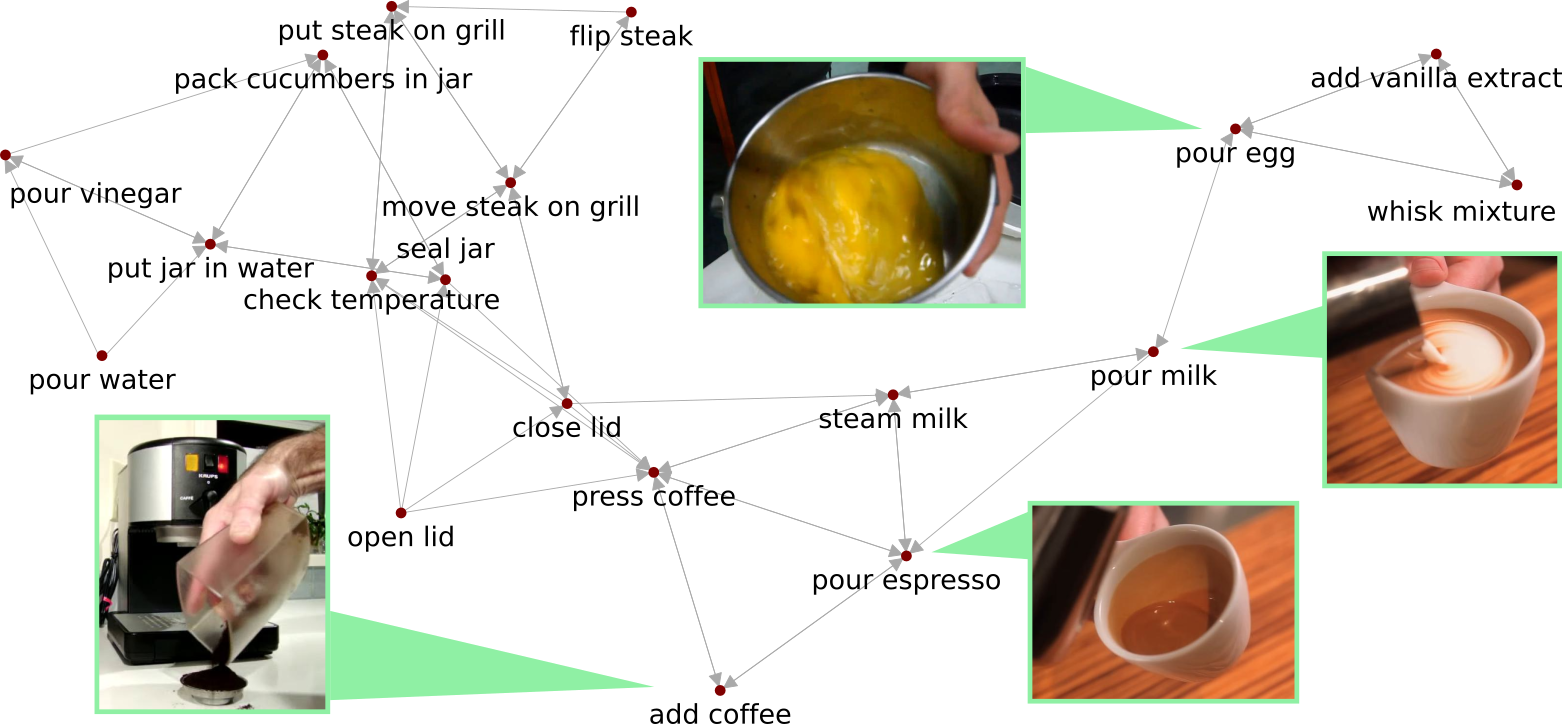}
 \vspace{0.5em}
 \caption{Procedural graph\protect\footnotemark $\mathbf{G}$ is a directed graph where nodes are steps of activity and edges are chains of steps that are mined from video data. The graph plays an important role in modeling procedures involved in multi-step tasks to train different instructional understanding objectives (\eg, \ref{eq:ar_graph}, \ref{eq:sr_graph}, \ref{eq:sa_graph}, \ref{eq:pa_graph}, \ref{eq:pap_graph}, with step and order mistake detection in \S\ref{subsec:mistake}) and create online streaming dialog.}
 \label{fig:proc_graph}
\end{figure}

In addition to the video clips and the query, we also consider a procedural graph $\mathbf{G}$ for generating the answer $\mathbf{A}$.
\begin{equation}
    p(\mathbf{A} | \keyword{\mathbf{V}}, \mathbf{Q}, \keywordtwo{\mathbf{G}}) = \prod_{i=0}^{L-1} p(x_i | \keyword{\mathbf{V}}, \mathbf{Q}, \keywordtwo{\mathbf{G}}, x_{<i})~\label{eq:MLLM_with_G}
\end{equation}%
\paragraph{Procedural Graph Construction}
Before the \textit{training} phase, we construct a procedural graph by mining the training data using Alg.~\ref{alg:graph_construction}. The graph $\mathbf{G} = (\mathcal{V}_{\mathbf{G}}, \mathcal{E}_{\mathbf{G}})$ consists of a vertex set $\mathcal{V}_{\mathbf{G}}$ and an edge set $\mathcal{E}_{\mathbf{G}}$.  We obtain the nodes in $\mathcal{V}_{\mathbf{G}}$ using the function $\mathrm{getAnn}(\cdot): \mathbf{V} \mapsto \mathcal{V}$  which gets the action annotation (\eg, add milk) $v_t$ for clips of $\mathbf{v}_t$ present in a task video $\mathbf{V}$ (\eg, how to make latte). The edges represent temporally ordered transitions  between two consecutive actions $(v_{t-1}, v_{t})$ observed in the task videos, which may be instructional~\cite{miech2019howto100m, tang2019coin, zhukov2019cross, afouras2024ht} or procedural~\cite{koupaee2018wikihow, puig2018virtualhome} in nature. An example subgraph in $\mathbf{G}$ is illustrated in Fig.~\ref{fig:proc_graph}.

\paragraph{Online Assistance}
During the \textit{inference} phase, we construct an online search path $\widehat{\mathbf{G}}_t$ as the video scene unfolds. For this, whenever we predict a change in action (using action recognition), we map it to a node in $\mathbf{G}$. Given the auto-regressive model recognizes action $\mathbf{a}_t$ as free-form text, we use a one-hot (similarity) mapping to select nodes in $\mathbf{G}$ and add it to a (predicted) online search path $\widehat{\mathbf{G}}$ (see Alg. \ref{alg:online_assist}).
\begin{align}
\text{node } \widehat{v}_t = \arg\max\big(\mathbbm{1}_{\mathcal{V}_{\mathbf{G}}}(\mathbf{a}_{t})\big) &, \quad \text{edge } (\widehat{v}_{t-1}, \widehat{v}_t) \nonumber\\
\keywordtwo{\widehat{\mathbf{G}}_t} = \Big(\underbrace{\{\widehat{v}_0\} + \{\widehat{v}_t\}}_{\keywordtwo{\mathcal{V}_{\widehat{\mathbf{G}}_t}}}, \underbrace{\big\{(\widehat{v}_{t-1}, \widehat{v}_t)\big\}}_{\keywordtwo{\mathcal{E}_{\widehat{\mathbf{G}}_t}}}\Big) &, \quad t \in (0, |\mathbf{V}|) \label{eq:graph}
\end{align}%
Projecting an online video onto a predicted subgraph $\widehat{\mathbf{G}} (\in \mathbf{G})$ enables the possibility of leveraging $\widehat{\mathbf{G}}$ alongside video and query embedding all the aforementioned tasks described in \S\ref{sec:prob_formu}. We hypothesize this reduces the burden of reasoning (needed for plan/action recognition and prediction) of the LLM by using plan prefixes $\widehat{\mathbf{G}}$ as part of the input.

The tasks defined in \S\ref{sec:prob_formu} can now be defined as
\begin{gather}\label{eq:ar_graph}
p(\mathbf{A} | \keyword{\mathbf{V}}, \mathbf{Q}_{\texttt{TR}}, \keywordtwo{\widehat{\mathbf{G}}_t}) \tag{TR} \\
p\bigg(\mathbf{a}_t | \keyword{\mathbf{v}_t}, \mathbf{Q}_{\texttt{AR}}, \underbrace{\Big(\{v_0\} + \{v_{\keyword{<t}}\}, \big\{(v_{\keyword{<t - 1}}, v_{\keyword{<t}})\big\}\Big)}_{\keywordtwo{\widehat{\mathbf{G}}}_{\keyword{<t}}}\bigg) \label{eq:sr_graph} \tag{AR} \\
p(\mathbf{a}_t | \keyword{\mathbf{v}_{\keyword{<t}}}, \mathbf{Q}_{\texttt{AP}}, \keywordtwo{\widehat{\mathbf{G}}}_{\keyword{<t}}) \label{eq:sa_graph} \tag{AP}
\end{gather}%
For \textit{Plan Prediction}, we choose to look at two variants-- with (PP+) and without (PP) knowing the target task $\mathbf{T}$:
\begin{gather}
p(\keyword{\mathbf{a}_{\geq t}} | \keyword{\mathbf{v}_{\keyword{<t}}}, \mathbf{Q}_{\texttt{PP}}, \keywordtwo{\widehat{\mathbf{G}}}_{\keyword{<t}}), \label{eq:pa_graph} \tag{PP} \\
p(\keyword{\mathbf{a}_{\geq t}} | \keyword{\mathbf{v}_{\keyword{<t}}}, \mathbf{Q}_{\texttt{PP}}, \keywordtri{\mathbf{T}}, \keywordtwo{\widehat{\mathbf{G}}}_{\keyword{<t}}) \tag{PP+}
\label{eq:pap_graph}
\end{gather}
where the number of procedural steps in $|\keyword{\mathbf{a}_{\geq t}}| > |\mathbf{a}_t| = 1$.

\subsection{Incorporating Conversational Context}\label{subsec:context}

\footnotetext{Code for \textit{visualization} is adapted from \href{https://github.com/facebookresearch/TaskGraph}{facebookresearch/TaskGraph}~\cite{ashutosh2024video}.}

To support streaming dialog with a user, previous works consider annotation efforts that require significant human effort~\cite{chen2024videollm} or train models using single-shot question answering~\cite{yu2019activitynet}. To overcome these limitations, our approach uses the procedural graph to naturally generate this type of annotation.
Both at \textit{training} and \textit{inference} time, a verbalization process is used for graph nodes. Specifically, we construct a conversational context by varying $t \in (0, |\mathbf{V}|)$ and use the conversation template:
\begin{equation}
\begin{aligned}
&\{\texttt{\small \textbf{Stream}:<} \keyword{\mathbf{v}_0}\texttt{\small >}; \texttt{\small \textbf{User}:<} \mathbf{Q}\texttt{\small >}; \texttt{\small \textbf{Assistant}:<}\keywordtwo{\widehat{v}_0}\texttt{\small >}; \dots; \\
&\texttt{\small \textbf{Stream}:<} \keyword{\mathbf{v}_t}\texttt{\small >}; \texttt{\small \textbf{User}:<} \mathbf{Q}\texttt{\small >}; \texttt{\small \textbf{Assistant}:<} \keywordtwo{\widehat{v}_t}\texttt{\small >}\}, \text{\small $0 < t < |\mathbf{V}|$} \notag
\end{aligned}
\end{equation}%
\textbf{Negative Choice Question.} We also augment the dialog guided by the procedure graph $\mathbf{G}$ by adding negative nodes that are distinct from $v_t$ and its neighboring nodes $\mathcal{N}(v_t)$. These negative nodes $\keywordone{v({\notin \mathcal{N}(v_t)\cup v_t}}$), enhance the ability of the model to distinguish between relevant and irrelevant information in the dialogue stream:
\begin{equation}
\begin{aligned}
\{\texttt{\small \textbf{Stream}:<}\keyword{\mathbf{v}_t}\texttt{\small >}; \texttt{\small \textbf{User}:<}\mathbf{Q}\texttt{\small >, should I <} \keywordone{v_{\notin \mathcal{N}(v_t)\cup v_t}}\texttt{\small >?}; \\
\texttt{\small \textbf{Assistant}: No, you should do <}\keywordtwo{v_t}\texttt{\small > instead.}\} \notag
\end{aligned}
\end{equation}

\noindent \textbf{Multiple Choice Question} is constructed via the template:
\begin{equation}
\begin{aligned}
\{&\texttt{\small \textbf{Stream}:<}\keyword{\mathbf{v}_t}\texttt{\small >}; \texttt{\small \textbf{User}:<}\mathbf{Q}\texttt{\small >, in one of <} \keywordtwo{v_{\in \mathcal{N}(v_{t-1})}}\texttt{\small >}; \\
&\texttt{\small \textbf{Assistant}: Yes, please do <}\keywordtwo{v_t}\texttt{\small >.}\} \notag
\end{aligned}
\end{equation}
where $v_t$ was added to $\mathcal{N}(v_{t-1})$ while constructing $\mathbf{G}$ (see L\ref{l:neighbor} in Alg.~\ref{alg:graph_construction}).

\input{algos/graph_construction}

\subsection{Error Detection}\label{subsec:mistake}

Beyond the five main tasks, also studied in prior works \cite{chen2024videollm}, our graph implementation allows us to perform well on two auxiliary tasks that leads to a more holistic validation of the instructional video understanding problem.

\paragraph{Incorrect Action Detection}
We create samples with incorrect actions by modifying each video in the data. Precisely, we randomly replace one step with an incorrect step $\keywordone{v_{\notin \mathcal{N}(v_t)\cup v_t}}$. This leads to an erroneous graph $\keywordone{\bar{\mathbf{G}}}$ generation. The task is to identify this mistaken step within the sequence and we measure the average accuracy of correctly identifying the index of the mistaken step:
\begin{equation}
\mathcal{V}_{\keywordone{\bar{\mathbf{G}}}} = \{v_0, \dots, \keywordone{v_{\notin \mathcal{N}(v_t)\cup v_t}}, \dots, v_{|\mathbf{V}| - 1}\}
\end{equation}%
\paragraph{Incorrect Order Detection}
By randomly shuffling the order of steps, we create a dataset for detecting mistakes in the ordering. The framework is then trained to determine whether the steps in a given video are in the correct order (or not). As one task can be achieved via different plans (or paths in the graph), we ensure the randomly shuffled order of actions is different from all action orderings present in the videos belonging to a particular task. Our evaluation metric is the average accuracy of the model in predicting whether a sequence is correctly ordered or not on the test split data:
\begin{equation}
\mathcal{E}_{\keywordone{\bar{\mathbf{G}}}} = \{\dots, (v_{t-1}, \keywordone{v_{\neq t}}), \dots\}
\end{equation}

\input{algos/online_graph}

\section{Experiments}%
\label{sec:expmt}
\subsection{Benchmarks and Metrics}

\input{tables/stats}

In our experiments, we consider two prominent video-based datasets-- COmprehensive INstructional video analysis (COIN; \cite{tang2019coin})  and CrossTask~\cite{zhukov2019cross}. These datasets encompass a wide range of everyday activities with explicitly defined steps, making them ideal for instructional video analysis. As highlighted in \autoref{tab:statistics}, COIN contains 10,166 videos covering 180 different activities and 746 distinct steps, organized in a three-level semantic structure of domain, activity, and step. It primarily focuses on daily tasks (cleaning, repairing, etc.) related to vehicles, gadgets, etc. CrossTask comprises 4,462 videos across 83 activities, covering tasks related to cooking, car maintenance, crafting, home repairs, etc.
The tasks and action annotations in CrossTask are derived from wikiHow~\cite{koupaee2018wikihow}.
Both datasets aim to establish a rich semantic taxonomy for organizing instructional videos.
We organize these datasets to obtain labeled data for all tasks described in \S\ref{sec:prob_formu}-- Task Recognition~\eqref{eq:ar_graph}, Action Recognition~\eqref{eq:sr_graph}, Action Prediction~\eqref{eq:sa_graph}, Plan Prediction~\eqref{eq:pa_graph} with a known goal~\eqref{eq:pap_graph}-- and \S\ref{subsec:mistake}-- Incorrect Action and Ordering detection. Precisely, we report the number of data samples, videos, tasks, actions, incorrect actions, and order-shuffled examples in Table~\ref{tab:statistics}. We use the same number of training samples across all methods and accuracy as the metric for evaluating performance on all the tasks.

\subsection{Implementation Details}

We employ CLIP-ViT-L-336~\cite{radford2021learning, dosovitskiy2021an} as the video frame encoder, a 2-layer MLP as the connector, and Mistral-7B-Instruct~\cite{jiang2023mistral} as the LLM. Each video frame is encoded into 10 tokens. Further, we use LoRA~\cite{hu2022lora} for training, applying it to all linear layers with a rank of 128 and a scaling factor of 256. With a batch size of 128 and gradient accumulation over 16 iterations, we observe a training time duration of $\approx$ 12 hours for 2 epochs when these runs are parallelized on 8 A100 GPUs on AWS' P4d instances. We now consider some of the design choices made for our model architecture that were made based on experimental results.

\input{tables/temporal_aggregator_table}

\paragraph{Temporal Aggregation Operations}
While various operations, such as flattening, averaging, or custom pooling, can be considered for aggregating along the temporal dimensions, we wanted to determine this empirically based performance of these alternatives of all the tasks in \S\ref{sec:prob_formu}. In \autoref{tab:temporal_operators}, we observe that Pooling yields the best results across all tasks. We hypothesize that Flatten, which simply concatenates features across the temporal dimension, introduces a lot of (irrelevant) data into the visual representation, making it difficult for the LLM to identify the needles in the visual haystack needed to excel on the tasks. On the other hand, averaging across all the temporal features risks losing out on fine-grained information that might be relevant to the downstream task. Spatial pooling strikes a good balance by reducing the spatial dimension at each time step, preventing the LLM from being overwhelmed with extra information. %

\input{tables/video_retrieval_table}

\paragraph{CLIP Backbone Selection}
To determine the best visual encoding for our model, we consider the precision, recall, and F1 metrics for a text-to-video retrieval task. In this task, we use a task name as the text query $\mathbf{Q}_{\texttt{TR}}$ and retrieve relevant videos. \autoref{tab:retrieval} highlights that the CLIP-H-14 backbone results in the best retrieval performance.

\paragraph{Effect of Dataset on Graph Construction}
To test the robustness of our graph construction approach on different data sources, in \autoref{tab:graph_source}, we observe the performance of using a $\mathbf{G}$ constructed from three different sources on the downstream tasks. We consider (1) the top-5 retrieved videos from text-to-video retrieval (Table~\ref{tab:retrieval}), (2) WikiHow~\cite{koupaee2018wikihow}, and (3) the entire training dataset as the three alternative data sources. On the COIN dataset, using retrieved videos for graph construction outperform others for~\ref{eq:sr_graph} (79.1\%) and~\ref{eq:pa_graph} (52.5\%), while using the entire training dataset excels on~\ref{eq:sa_graph} (65.9\%), ~\ref{eq:pa_graph} (59.1\%) and~\ref{eq:ar_graph} (98.9\%). For CrossTask, using the entire training set results in the best performance across all tasks except on~\ref{eq:sr_graph} (70.1\% < 71.5\%). We note that $\mathbf{G}$ constructed with WikiHow consistently underperforms across all tasks in both datasets.
We postulate that the comparable performance of Retrieved on recognition-related tasks can be attributed to their more focused contextual scope, enabling more precise information extraction. Conversely, the full training dataset excels in future action/plan anticipation tasks due to a more holistic understanding of action dependencies in tasks gathered from a larger and diverse set of task videos.

\subsection{Procedural Graph Usage for Inference}\label{subsec:infer_graph}

After our procedural graph extraction, we can leverage it at inference time regardless of the Multi-modal LLM (MLLM) used for the online assistance tasks. In \autoref{tab:improvement}, we highlight that considering the online graph path construction and incorporating it as input can unanimously improve the performance of any MLLM across all tasks (in \S\ref{sec:prob_formu}) and datasets. For the VideoLLM-online model~\cite{chen2024videollm} on the COIN dataset, the addition of our graph implementation ($\keyword{\mathbf{V}}\mathbf{Q}\keywordtwo{\mathbf{G}}$) led to substantial improvements (notably, absolute gains of +8.2\% on \ref{eq:sr_graph}, +13.7 on \ref{eq:sa_graph}, and +8.7 on \ref{eq:pap_graph}) even when the baseline ($\keyword{\mathbf{V}}\mathbf{Q}$) used an enhanced version of Llama-3-\textbf{8B}.
The GPT models~\cite{achiam2023gpt} also benefited significantly from our approach. When augmented with our graph implementation, GPT-4o-mini showed improvements ranging from +2.5  to +16.9 percentage points across various tasks on both datasets. Further, GPT-4-turbo and GPT-4o models also exhibited substantial improvements when integrated with our graph approach; notably, absolute gains of +21.7\% and +17.8\% on \ref{eq:sa_graph} for the two models respectively. These results reinforce that augmenting dependencies explicitly (via our graph approach) instead of heavily relying on the planning capabilities of LLMs in a multi-modal setting can improve task performance. We now show that leveraging the procedural task graphs for multi-task learning can provide further gains.

\input{tables/data_sources}

\input{tables/comparisons}

\input{tables/improvements}

\subsection{Efficacy of \alname}

In \autoref{tab:sota}A, we provide a comprehensive comparison of our method \alname~against various State-of-The-Art (SoTA) approaches for instructional video understanding on COIN. While, the \alname~base model ($\keyword{\mathbf{V}}\mathbf{Q}$) achieves the second-best scores on action recognition (\ref{eq:sr_graph}) and prediction (\ref{eq:sa_graph}) beating existing baselines, the latest work on VideoLLM-online+ \cite{chen2024videollm} shows second best performance on task recognition and plan prediction tasks that need longer dependency resolution. We hypothesize that this is due to the better reasoning capabilities of the LLM backbone for VideoLLM-online+ (i.e. LLama-3-8B) compared to our LLM backbone (i.e. Mistral-7B-Instruct). Regardless, when we incorporate our graph-based component ($\keyword{\mathbf{V}}\mathbf{Q}\keywordtwo{\mathbf{G}}$), \alname's performance improves exceptionally across all tasks beating all previous baselines (and the graph-augmented approaches developed in \S\ref{subsec:infer_graph}) across all existing video assistance tasks described in \S\ref{sec:prob_formu}. Notably, \alname~improves \ref{eq:ar_graph}~performance to $98.9\%$ and all action and plan prediction tasks outperforming the previous best LLM-based method, VideoLLM-online+. We hypothesize that VideoLLM-online+, which creates simple augmented questions and transfers the burden of reasoning over the procedural video data to the LLM, is worse off than \alname's better procedural context understanding due to the incorporation of task-graphs.

In \autoref{tab:sota}B, we show similar results on CrossTask. While traditional video understanding models, such as S3D and SlowFast, achieve respectable performance (45.3\% and 48.5\% respectively) on \ref{eq:sr_graph}, newer models, like VideoCLIP and TimeSformer, improve further (reaching~60.1\% and 60.9\% respectively). More recent, specialized models, such as DistantSup and TaskGraph (the latter uses graphs), push the accuracy further achieving 64.2\% and 64.5\% on ~\ref{eq:sr_graph}. We also consider proprietary API-based MLLMs (OpenAI variants) that show competitive performance, particularly on~\ref{eq:pa_graph} and~\ref{eq:pap_graph} tasks. \alname, when using only visual and query inputs, outperforms all previous methods across most tasks, achieving 65.1\% on~\ref{eq:sr_graph} and 97.6\% on~\ref{eq:ar_graph}. When we incorporate the graph-based approach, \alname~sets the SoTA performance with substantial improvement-- 70.1\% on~\ref{eq:sr_graph}, 49.7\% on~\ref{eq:sa_graph}, and nearly perfect 99.6\% score on~\ref{eq:ar_graph}. While we also perform the best on plan prediction, undoubtedly the most difficult task, our numbers (39.0\% on~\ref{eq:pa_graph} and 42.9\% on~\ref{eq:pap_graph}) highlight a large scope for improvement.

\paragraph{Error Detection}
On the auxiliary tasks for detecting errors in action and ordering (\S\ref{subsec:mistake}), we maintain the same train/test splits across methods and report the average accuracy of correctly identifying action and ordering errors in \autoref{tab:sota}C. \alname~demonstrates a significant improvement over all baselines when, esp. when using the procedural-graph implementation boosing error detection for action from 40.9\% to 51.6\%. For incorrect order detection, we observe a smaller improvement from 42.1\% to 44.1\%. While having procedural graphs helps unanimously, we hypothesize that it is particularly effective in capturing short-term dependency relationships between actions, allowing to better identify out-of-context or incorrect steps. This hypothesis also explains (albeit \textit{post-facto}) the lower magnitude of improvements seen for plan prediction tasks compared to action prediction tasks when $\mathbf{G}$ is incorporated in the earlier experiments.

\begin{figure*}[t]
 \centering
 \includegraphics[width=\textwidth]{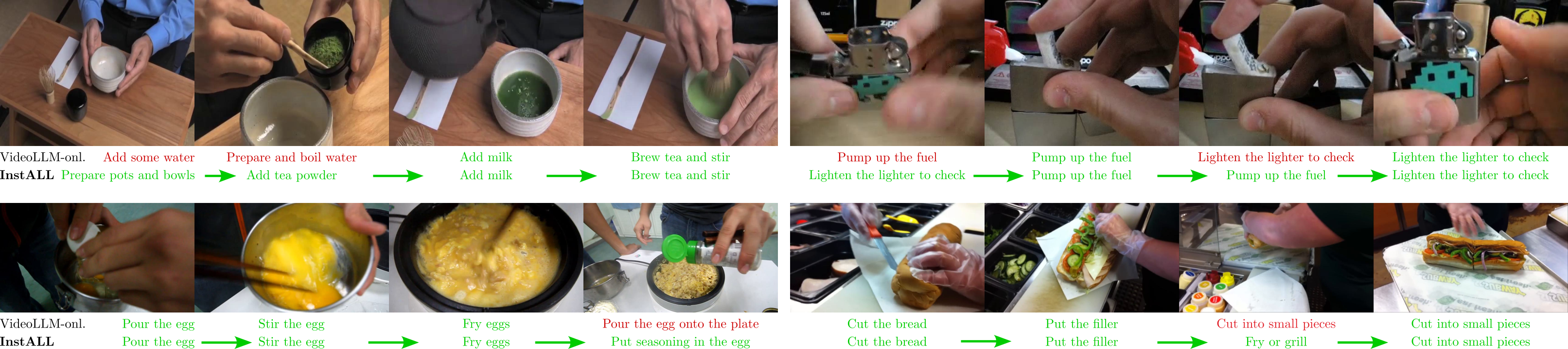}
 \vspace{0.5em}
 \caption{Qualitative comparison of our graph-based \alname~versus VideoLLM-online~\cite{chen2024videollm} for~\eqref{eq:sr_graph}. Our \alname~is aware of the position of the step in the entire procedure (\keywordtwo{arrows}) and predicts the steps accurately. At the same time, VideoLLM-online misinterprets the order by relying solely on visual cues. \keywordone{Red} texts denote incorrect steps while \keywordtwo{green} texts denote correct steps. \textbf{Best viewed in color.}}
 \label{fig:quantitative}
\end{figure*}

\paragraph{Qualitative Comparison}
To illustrate the effectiveness of \alname, we present four qualitative comparisons in Fig.~\ref{fig:quantitative}, which showcases the procedure for making matcha tea, refilling a lighter, cooking omelet, and making a sandwich. Procedures often involve repetitive steps and can present ambiguous visual information, making challenges for purely visual systems. VideoLLM-online~\cite{chen2024videollm} (top), which relies solely on visual representations and piggybacks on LLM's reasoning capabilities, without the benefit of our graph-based approach, misinterprets the step order. In contrast, our method models this procedure as a graph and injects it into the LLM, enables accurate prediction of the step order. Incorporating $\mathbf{G}$ provides crucial context and relational information, relieving reasoning expectations on the LLM, and helping the model correctly interpret and predict the sequence of steps, even in scenarios where visual cues alone might be ambiguous.

\section{Conclusion}

In this paper, we presented a novel approach \alname~for instructional video understanding. \alname~leverages graph-based representations in conjunction with visual and textual embedding for adapter-style Multi-modal LLMs (MLLMs). \alname~demonstrates significant improvements across a wide range of tasks including Action Recognition, Action Prediction, Plan Prediction, Task Recognition, and Error Identification. Injecting procedural task knowledge as graphs into the LLMs, we provide an accurate and rich representation of complex, multi-step processes, easing the reasoning burden on the LLMs. Extensive experiments showcase the consistent superiority of \alname~compared to a wide range of approaches, ranging from traditional video understanding models and recent LLM-based approaches. Overall, our work contributes to building a solution that achieves SoTA across all tasks which is key for enabling visually aware assistants for procedural task videos.

\paragraph{Future Works}
While the online graph yields unanimous improvements, we observed, similar to prior work \cite{roy-etal-2024-flap}, that it can result in prediction errors as the model cannot faithfully follow the dependencies in the graph. We note that such errors compound across prediction steps, limiting their efficacy on plan prediction. We believe methods that improve ways to incorporate structured knowledge and improve the reasoning abilities of LLMs will be the path forward.

%% file: tables/all_related_works_table.tex
\begin{table}[t]
\footnotesize
\centering
\setlength\tabcolsep{4pt}
\begin{tabular}{lcccccccc}
\toprule
 \multicolumn{1}{c}{\multirow{2}{*}{\textbf{Method}}} & \multicolumn{1}{c}{\multirow{2}{*}{\textbf{Onl.}}} & \multicolumn{1}{c}{\multirow{2}{*}{\textbf{QA}}} & \multicolumn{1}{c}{\multirow{2}{*}{\textbf{PG}}} & \multicolumn{5}{c}{\textbf{Tasks}} \\ \cmidrule{5-9}
 & & & & Retr. & TR & AR & AP & PP \\
\midrule
ClipBERT~\cite{lei2021less} & \xmark & \xmark & \xmark & \xmark & \cmark & \cmark & \xmark & \xmark \\
ProcedureVRL~\cite{zhong2023learning} & \xmark & \xmark & \xmark & \xmark & \cmark & \cmark & \cmark & \xmark \\
TimeSformer~\cite{bertasius2021space} & \xmark & \xmark & \xmark & \xmark & \cmark & \cmark & \cmark & \cmark \\
DistantSup~\cite{lin2022learning} & \xmark & \xmark & \xmark & \xmark & \cmark & \cmark & \cmark & \cmark \\
VideoTF~\cite{narasimhan2023learning} & \xmark & \xmark & \xmark & \xmark & \cmark & \cmark & \cmark & \cmark \\
Paprika~\cite{zhou2023procedure} & \xmark & \xmark & \cmark & \xmark & \cmark & \cmark & \cmark & \xmark \\
TaskGraph~\cite{ashutosh2024video} & \xmark & \xmark & \cmark & \xmark & \cmark & \cmark & \cmark & \xmark \\
\keyword{\textbf{VideoLLM-onl.}}~\cite{chen2024videollm} & \cmark & \cmark & \xmark & \xmark & \cmark & \cmark & \cmark & \cmark \\ \midrule
\keyword{\textbf{\alname} (Ours)} & \cmark & \cmark & \cmark & \cmark & \cmark & \cmark & \cmark & \cmark \\
\bottomrule
\end{tabular}
\vspace{0.5em}
\caption{Our work considers an online (\textbf{Onl.}), conversational (\textbf{QA}) setting for procedural tasks, where we can leverage Procedural Graphs (\textbf{PG}). In the context of online assistance, our approach can support offline text-to-video retrieval (\textbf{Retr.}), and all sub-tasks proposed in prior works, such as Video Task Recognition (TR), Action Recognition (AR), and Action Prediction (AP), Plan Prediction (PP) (\S\ref{sec:prob_formu}). In addition, we formulate two new auxiliary tasks related to error detection that are important for effective assistance (see \S\ref{subsec:mistake}). Methods in \keyword{\textbf{blue}} use Multi-modal LLMs (MLLMs).}\label{tab:features}
\end{table}

%% file: tables/task_coverage_table.tex
\begin{table}[t]
\footnotesize
\centering
\setlength\tabcolsep{4pt}
\begin{tabular}{lcccc}
\toprule
 \textbf{Method} & \textbf{\#Anns} & \textbf{Multi-choice} & \textbf{Negation} & \textbf{Mistake} \\
\midrule
VideoLLM-onl.~\cite{chen2024videollm} & 134K & \xmark & \xmark  & \xmark \\ \midrule
\textbf{\alname} (Ours) & \textbf{245K} & \cmark & \cmark & \cmark \\
\bottomrule
\end{tabular}
\vspace{0.5em}
\caption{Num. of annotated samples and tasks featured in \alname.}\label{tab:prompt_types}
\end{table}

%% file: algos/graph_construction.tex
\begin{figure}[t]
\small
\vspace{-\baselineskip}
\begin{algorithm}[H] %
\caption{Procedural Graph $\mathbf{G}$ Construction}
\label{alg:graph_construction}
\begin{algorithmic}[1]
\INPUT{$\mathcal{V}_{\mathbf{G}} \gets \varnothing$, $\mathcal{E}_{\mathbf{G}} \gets \varnothing$}
\FORALL{videos $\mathbf{V}$}
\STATE $\mathcal{V}_{\mathbf{G}}\mathrm{.insert}\big(\mathrm{getAnn}(\mathbf{v}_0)\big)$ \COMMENT{ get category of step clip}
\FOR{$\mathbf{v}_t \in \{\mathbf{v}_1, \dots, \mathbf{v}_{|\mathbf{V}| - 1}\}$}
\STATE $v_{t-1}, v_{t} \gets \mathrm{getAnn}(\mathbf{v}_{t-1}), \mathrm{getAnn}(\mathbf{v}_{t})$\label{l:getAnn}
\STATE \textbf{if $v_{t} \notin \mathcal{V}_{\mathbf{G}}$ then} $\mathcal{V}_{\mathbf{G}}\mathrm{.insert}(v_{t})$
\STATE $\mathcal{E}_{\mathbf{G}}\mathrm{.insert}(v_{t-1}, v_{t})$\label{l:neighbor}
\ENDFOR
\ENDFOR
\STATE $\mathbf{G} \gets (\mathcal{V}_{\mathbf{G}}, \mathcal{E}_{\mathbf{G}})$
\end{algorithmic}
\end{algorithm}
\vspace{-1.5\baselineskip}
\end{figure}

%% file: algos/online_graph.tex
\begin{figure}[t]
\small
\vspace{-\baselineskip}
\begin{algorithm}[H] %
\caption{Online Assistance w/ Procedural Graph}
\label{alg:online_assist}
\begin{algorithmic}[1]
\INPUT{Video $\mathbf{V}$, $\mathcal{V}_{\widehat{\mathbf{G}}} \gets \varnothing$, $\mathcal{E}_{\widehat{\mathbf{G}}} \gets \varnothing$, $\widehat{\mathbf{G}} \gets (\mathcal{V}_{\widehat{\mathbf{G}}}, \mathcal{E}_{\widehat{\mathbf{G}}})$}
\FOR{clip $\mathbf{v}_t \in \{\mathbf{v}_0, \dots, \mathbf{v}_{|\mathbf{V}| - 1}\}$}
\STATE $\mathbf{A} \gets p\big(\mathbf{A} | \mathbf{V}, \mathbf{Q}, \widehat{\mathbf{G}}\big)$
\STATE $\widehat{v}_t \gets \arg\max\big(\mathbbm{1}_{\mathcal{V}_{\mathbf{G}}}(\mathbf{A})\big)$
\STATE $\mathcal{V}_{\widehat{\mathbf{G}}}\mathrm{.insert}(\widehat{v}_t)$
\STATE \textbf{if $t > 0$ then} $\mathcal{E}_{\widehat{\mathbf{G}}}\mathrm{.insert}(\widehat{v}_{t-1}, \widehat{v}_t)$
\STATE $\widehat{\mathbf{G}} \gets (\mathcal{V}_{\widehat{\mathbf{G}}}, \mathcal{E}_{\widehat{\mathbf{G}}})$
\ENDFOR
\end{algorithmic}
\end{algorithm}
\vspace{-1.5\baselineskip}
\end{figure}

%% file: tables/stats.tex
\begin{table*}[!t]
\resizebox{\textwidth}{!}{
\setlength\tabcolsep{6pt}
{\begin{tabular}{lcccccccccc}
\toprule
& \textbf{\#TR} & \textbf{\#AR} & \textbf{\#AP} & \textbf{\#PP} & \textbf{\#PP+} & \textbf{\#Action}\keywordone{$_{\bar{\mathbf{G}}}$} & \textbf{\#Order}\keywordone{$_{\bar{\mathbf{G}}}$} & \textbf{\#Videos} & \textbf{\#Tasks$_{\neq}$} & \textbf{\#Actions$_{\neq}$} \\ \midrule
CrossTask~\cite{zhukov2019cross} & 4,462 & 36,117 & 31,656 & 27,253 & 27,253 & 4,462 & 4,462 & 4,462 & 83 & 105 \\
COIN~\cite{tang2019coin} & 10,166 & 39,747 & 29,584 & 19,420 & 19,420 & 10,166 & 10,166 & 10,166 & 180 & 746 \\
\bottomrule
\end{tabular}}
}
\vspace{0.5em}
\caption{Statistics of samples for the constructed tasks \ref{eq:ar_graph}, \ref{eq:sr_graph}, \ref{eq:sa_graph}, \ref{eq:pa_graph}, and \ref{eq:pap_graph} from video datasets and our graph. \textbf{\#Tasks$_{\neq}$} and \textbf{\#Actions$_{\neq}$} denote the number of \textbf{\textit{unique}} activity and step categories. Training and testing splits are roughly 80:20 with no videos in common.}
\label{tab:statistics}
\end{table*}

%% file: tables/temporal_aggregator_table.tex
\begin{table}[t!]
    \small
    \centering
    \begin{tabular}{lcccccc}
    \toprule
    & \multicolumn{1}{c}{\textbf{Shape}} & \textbf{AR} & \textbf{AP} & \textbf{PP} & \textbf{PP+} & \textbf{TR} \\ \midrule
    Flatten & $T\!\!\times\!\! N\!\!\times\!\! D$ & 54.0 & 47.1 & 47.3 & 52.3 & 77.3 \\
    Averag. & $1\!\!\times\!\! N\!\!\times\!\! D$ & 60.8 & 58.0 & 49.5 & 57.4 & 79.4 \\
    Pooling & $T\!\!\times\!\! N\!\!\times\!\! D/W$ & \textbf{78.5} & \textbf{65.9} & \textbf{52.3} & \textbf{59.1} & \textbf{98.9} \\
    \bottomrule
    \end{tabular}
    \vspace{0.5em}
    \caption{Performance of Temporal Aggregation methods.}
    \label{tab:temporal_operators}
\end{table}

%% file: tables/video_retrieval_table.tex
\begin{table}[t!]
    \centering
    \small
    \begin{tabular}{lccc}
    \toprule
    & \textbf{Recall} & \textbf{Precision} & \textbf{F1-Score} \\ \midrule
    CLIP-B-16 & 65.1\% & 68.4\% & 66.7\% \\
    CLIP-L-14 & 78.3\% & 89.3\% & 83.4\% \\
    CLIP-H-14 & \textbf{92.2}\% & \textbf{95.8}\% & \textbf{93.8}\% \\
    \bottomrule
    \end{tabular}
    \vspace{0.5em}
    \caption{Retrieval performance of VectorRAG alternatives.}
    \label{tab:retrieval}
\end{table}

%% file: tables/data_sources.tex
\begin{table*}[!t]
\begin{minipage}{.495\textwidth}
\resizebox{\textwidth}{!}{
\setlength\tabcolsep{12pt}
{\begin{tabular}{lcccccc}
\toprule
\multicolumn{1}{c}{\multirow{2}{*}{$\keywordtwo{\mathbf{G}}$}} & \multicolumn{5}{c}{\textbf{COIN}} \\ \cmidrule(lr){2-6}
& AR & AP & PP & PP+ & TR \\ \midrule
Retrieved & \textbf{79.1} & 63.8 & \textbf{52.5} & 57.3 & 98.8 \\ 
WikiHow & 63.6 & 48.2 & 44.6 & 46.4 & 85.9 \\ 
Trainset & 78.5 & \textbf{65.9} & 52.3 & \textbf{59.1} & \textbf{98.9} \\ 
\bottomrule
\end{tabular}}
}
\end{minipage} \hfill
\begin{minipage}{.495\textwidth}
\resizebox{\textwidth}{!}{
\setlength\tabcolsep{12pt}
{\begin{tabular}{lcccccc}
\toprule
\multicolumn{1}{c}{\multirow{2}{*}{$\keywordtwo{\mathbf{G}}$}} & \multicolumn{5}{c}{\textbf{CrossTask}} \\ \cmidrule(lr){2-6}
& AR & AP & PP & PP+ & TR \\ \midrule
Retrieved & \textbf{71.5} & 49.2 & 38.5 & 40.4 & \textbf{99.6} \\ 
WikiHow & 58.5 & 27.6 & 24.3 & 25.8 & 94.0 \\ 
Trainset & 70.1 & \textbf{49.7} & \textbf{39.0} & \textbf{42.9} & \textbf{99.6} \\ 
\bottomrule
\end{tabular}}
}
\end{minipage}
\vspace{0.5em}
\caption{Performance on different data sources for graph construction, including the retrieval results, WikiHow, and entire training dataset.} \label{tab:graph_source}
\end{table*}

%% file: tables/comparisons.tex
\begin{table*}[!t]
\small
 \begin{minipage}{.624\textwidth}
 \resizebox{\textwidth}{!}{
 \setlength\tabcolsep{10.65pt}
 {\begin{tabular}{lccccc}
 \toprule
 \multicolumn{1}{c}{\multirow{2}{*}{\textbf{A. Understanding}}} & \multicolumn{5}{c}{\textbf{COIN}} \\ \cmidrule(lr){2-6}
 & AR & AP & PP & PP+ & TR \\ \midrule
 ClipBERT~\cite{lei2021less} & 30.8 & \xmark & \xmark & \xmark  &65.4 \\
 TSN~\cite{wang2016temporal} & 36.5 & \xmark & \xmark & \xmark  &73.4 \\ \midrule
 S3D~\cite{xie2018rethinking} & 37.3 & 28.1 & -- & --  &70.2 \\ 
 SlowFast~\cite{feichtenhofer2019slowfast} & 39.6 & 25.6 & -- & --  &71.6 \\ 
 MIL-NCE~\cite{miech2020end} & 42.0 & 36.6 & -- & --  &76.6 \\ 
 VSM~\cite{zhou2023procedure} & 44.4 & 39.3 & -- & --  &82.2 \\ 
 Paprika~\cite{zhou2023procedure} & 51.0 & 43.2 & -- & --  &85.8 \\ 
 VideoCLIP~\cite{xu-etal-2021-videoclip} & 51.2 & 34.6 & -- & --  &72.5 \\ 
 ProcedureVRL~\cite{zhong2023learning} & 56.9 & 46.8 & -- & --  &90.8 \\ 
 TaskGraph~\cite{ashutosh2024video} & 57.2 & 40.2 & -- & --  &90.5 \\ 
 DistantSup~\cite{lin2022learning} & 54.1 & 39.4 & -- & 41.3  &90.0 \\ 
 TimeSformer~\cite{bertasius2021space} & 46.5 & 34.0 & 17.0 & 40.1  &85.3 \\ 
 VideoTF~\cite{narasimhan2023learning} & 56.5 & 42.4 & 40.2 & 46.4  &91.0 \\ 
 VideoLLM-online~\cite{chen2024videollm} & 59.8 & 48.1 & 47.9 & 52.9  &92.1 \\ 
 VideoLLM-online+~\cite{chen2024videollm} & 63.1 & 49.1 & \underline{49.8}& \underline{54.1}&\underline{92.7}\\ \midrule 
 GPT-4o-mini & 42.5 & 31.2 & 20.8 & 29.4& 64.2 \\ 
 GPT-4-turbo & 52.4 & 41.2 & 26.2 & 34.4 &68.6 \\ 
 GPT-4o & 64.7 & 43.6 & 33.0 & 41.8 & 69.9 \\ \midrule
 \textbf{\alname} ($\keyword{\mathbf{V}}\mathbf{Q}$) & \underline{70.2}& \underline{52.1}& 48.5& 50.3&89.5\\ 
 \textbf{\alname} ($\keyword{\mathbf{V}}\mathbf{Q}\keywordtwo{\mathbf{G}}$) & \textbf{78.5} & \textbf{65.9} & \textbf{52.3} & \textbf{59.1}  & \textbf{98.9} \\
 \bottomrule
 \end{tabular}}
 }
 \end{minipage} \hfill
 \begin{minipage}{.3795\textwidth}
 \resizebox{\textwidth}{!}{
 \setlength\tabcolsep{4pt}
 {\begin{tabular}{lcccccl}
 \toprule
 \multicolumn{1}{c}{\multirow{2}{*}{\textbf{B. Understanding}}} & \multicolumn{5}{c}{\textbf{CrossTask}} \\ \cmidrule(lr){2-6}
 & AR & AP & PP & PP+ & TR \\ \midrule
 S3D~\cite{xie2018rethinking} & 45.3 & 21.7 & -- & -- & 87.8 \\ 
 SlowFast~\cite{feichtenhofer2019slowfast} & 48.5 & 24.0 & -- & -- & 89.8 \\ 
 VSM~\cite{zhou2023procedure} & 58.9 & 57.9 & -- & -- & 62.2 \\ 
 MIL-NCE~\cite{miech2020end} & 59.9 & 58.0 & -- & -- & 61.7 \\ 
 VideoCLIP~\cite{xu-etal-2021-videoclip} & 60.1 & 26.0 & -- & -- & 92.3 \\ 
 TimeSformer~\cite{bertasius2021space} & 60.9 & 27.1 & -- & -- & 93.8 \\ 
 DistantSup~\cite{lin2022learning} & 64.2 & 29.7 & -- & -- & 95.2 \\ 
 TaskGraph~\cite{ashutosh2024video} & 64.5 & 30.2 & -- & -- & 96.0 \\ \midrule 
 GPT-4o-mini & 48.8 & 23.5 & 21.6 & 25.7 & 52.7 \\ 
 GPT-4-turbo & 51.5 & 27.2 & 20.4 & 26.4 & 63.6 \\ 
 GPT-4o & 52.9 & 35.0 & 25.7 & 33.2 & 60.8 \\ \midrule
 \textbf{\alname} ($\keyword{\mathbf{V}}\mathbf{Q}$) & 65.1 & 31.5 & 27.9 & 29.7 & 97.6 \\
 \textbf{\alname} ($\keyword{\mathbf{V}}\mathbf{Q}\keywordtwo{\mathbf{G}}$) & \textbf{70.1} & \textbf{49.7} & \textbf{39.0} & \textbf{42.9} & \textbf{99.6} \\
 \bottomrule
 \end{tabular}}
 }
 \resizebox{\textwidth}{!}{
 \setlength\tabcolsep{13pt}
 {\begin{tabular}{lcc}
 \toprule
 \multicolumn{1}{c}{\multirow{2}{*}{\textbf{C. Error Detection}}} & \multicolumn{2}{c}{\textbf{COIN}} \\ \cmidrule(lr){2-3}
 & Action & Order \\ \midrule
 SlowFast~\cite{feichtenhofer2019slowfast} & 28.6 & 26.1 \\
 MPNet~\cite{song2020mpnet} & 34.2 & 33.4 \\
 TimeSformer~\cite{bertasius2021space} & 37.6 & 31.8 \\
 VideoTF~\cite{narasimhan2023learning} & 41.7 & 35.4 \\ \midrule
 \textbf{\alname} ($\keyword{\mathbf{V}}\mathbf{Q}$) & 40.9 & 42.1 \\
 \textbf{\alname} ($\keyword{\mathbf{V}}\mathbf{Q}\keywordtwo{\mathbf{G}}$) & \textbf{51.6} & \textbf{44.1} \\
 \bottomrule
 \end{tabular}}
 }
 \end{minipage}
 \vspace{0.5em}
 \caption{Comparison against State-of-the-Art Instructional Video Understanding methods. (A) and (B) report performances of \ref{eq:sr_graph}, \ref{eq:sa_graph}, \ref{eq:pa_graph}, \ref{eq:pap_graph}, and \ref{eq:ar_graph} tasks on COIN~\cite{tang2019coin} and CrossTask~\cite{zhukov2019cross}, respectively. (C) reports the mistake detection in both step and order on COIN~\cite{tang2019coin}.} \label{tab:sota}
\end{table*}

%% file: tables/improvements.tex
\begin{table*}[!b]
\small
 \caption{Our procedural graph modeling improves overall performances of all \ref{eq:sr_graph}, \ref{eq:sa_graph}, \ref{eq:pa_graph}, \ref{eq:pap_graph}, and \ref{eq:ar_graph} tasks for LLM-based approaches.}
 \resizebox{\textwidth}{!}{
 \setlength\tabcolsep{4pt}
 {\begin{tabular}{lllllllllll}
 \toprule
 & \multicolumn{5}{c}{\textbf{COIN}} & \multicolumn{5}{c}{\textbf{CrossTask}} \\ \cmidrule(lr){2-6} \cmidrule(lr){7-11}
 & \multicolumn{1}{c}{AR} & \multicolumn{1}{c}{AP} & \multicolumn{1}{c}{PP} & \multicolumn{1}{c}{PP+} & \multicolumn{1}{c}{TR} & \multicolumn{1}{c}{AR} & \multicolumn{1}{c}{AP} & \multicolumn{1}{c}{PP} & \multicolumn{1}{c}{PP+} & \multicolumn{1}{c}{TR} \\ \midrule
 VideoLLM-online ($\keyword{\mathbf{V}}\mathbf{Q}$) & 59.8 & 48.1 & 47.9 & 52.9 & 92.1 & \multicolumn{1}{c}{--} & \multicolumn{1}{c}{--} & \multicolumn{1}{c}{--} & \multicolumn{1}{c}{--} & \multicolumn{1}{c}{--} \\ 
 VideoLLM-online+ ($\keyword{\mathbf{V}}\mathbf{Q}$) & 63.1\textsubscript{(+3.3)} & 49.1\textsubscript{(+1.0)} & 49.8\textsubscript{(+1.9)} & 54.1\textsubscript{(+1.2)} & 92.7\textsubscript{(+0.6)} & \multicolumn{1}{c}{--} & \multicolumn{1}{c}{--} & \multicolumn{1}{c}{--} & \multicolumn{1}{c}{--} & \multicolumn{1}{c}{--} \\
 VideoLLM-online+ ($\keyword{\mathbf{V}}\mathbf{Q}\keywordtwo{\mathbf{G}}$) & 71.3\textsubscript{(\keywordtwo{+8.2})} & 62.8\textsubscript{(\keywordtwo{+13.7})} & 53.5\textsubscript{(\keywordtwo{+3.7})} & 62.8\textsubscript{(\keywordtwo{+8.7})} & 95.3\textsubscript{(\keywordtwo{+2.6})} & \multicolumn{1}{c}{--} & \multicolumn{1}{c}{--} & \multicolumn{1}{c}{--} & \multicolumn{1}{c}{--} & \multicolumn{1}{c}{--} \\ \midrule
 GPT-4o-mini ($\keyword{\mathbf{V}}\mathbf{Q}$) & 42.5 & 31.2 & 20.8 & 29.4 & 64.2 & 48.8 & 23.5 & 21.6 & 25.7 & 52.7 \\ 
 GPT-4o-mini ($\keyword{\mathbf{V}}\mathbf{Q}\keywordtwo{\mathbf{G}}$) & 51.4\textsubscript{(\keywordtwo{+8.9})} & 48.1\textsubscript{(\keywordtwo{+16.9})} & 23.7\textsubscript{(\keywordtwo{+2.9})} & 36.2\textsubscript{(\keywordtwo{+6.8})} & 66.9\textsubscript{(\keywordtwo{+2.7})} & 62.0\textsubscript{(\keywordtwo{+13.2})} & 36.5\textsubscript{(\keywordtwo{+13.0})} & 24.5\textsubscript{(\keywordtwo{+2.9})} & 30.9\textsubscript{(\keywordtwo{+5.2})} & 72.1\textsubscript{(\keywordtwo{+19.4})}\\ \midrule
 GPT-4-turbo ($\keyword{\mathbf{V}}\mathbf{Q}$) & 52.4 & 41.2 & 26.2 & 34.4 & 68.6 & 51.5 & 27.2 & 20.4 & 26.4 & 63.6\\ 
 GPT-4-turbo ($\keyword{\mathbf{V}}\mathbf{Q}\keywordtwo{\mathbf{G}}$) & 60.5\textsubscript{(\keywordtwo{+8.1})} & 57.3\textsubscript{(\keywordtwo{+16.1})} & 29.4\textsubscript{(\keywordtwo{+3.2})} & 40.6\textsubscript{(\keywordtwo{+6.2})} & 72.0\textsubscript{(\keywordtwo{+3.4})} & 60.5\textsubscript{(\keywordtwo{+9.0})} & 48.9\textsubscript{(\keywordtwo{+21.7})} & 24.8\textsubscript{(\keywordtwo{+4.4})} & 29.2\textsubscript{(\keywordtwo{+2.8})} & 69.3\textsubscript{(\keywordtwo{+5.7})} \\ \midrule
 GPT-4o ($\keyword{\mathbf{V}}\mathbf{Q}$) & 64.7 & 43.6 & 33.0 & 41.8 & 69.9 & 52.9 & 35.0 & 25.7 & 33.2 & 60.8 \\ 
 GPT-4o ($\keyword{\mathbf{V}}\mathbf{Q}\keywordtwo{\mathbf{G}}$) & 71.9\textsubscript{(\keywordtwo{+7.2})} & 61.4\textsubscript{(\keywordtwo{+17.8})} & 35.7\textsubscript{(\keywordtwo{+2.7})} & 45.8\textsubscript{(\keywordtwo{+4.0})} & 76.5\textsubscript{(\keywordtwo{+6.6})} & 64.7\textsubscript{(\keywordtwo{+11.8})} & 42.1\textsubscript{(\keywordtwo{+7.1})} & 28.2\textsubscript{(\keywordtwo{+2.5})} & 38.2\textsubscript{(\keywordtwo{+5.0})} & 72.9\textsubscript{(\keywordtwo{+12.1})} \\
 \bottomrule
 \end{tabular}}
 \label{tab:improvement}
 }
\end{table*}